\documentclass[11pt]{article}
\usepackage{acl2016}
\usepackage{times}
\usepackage{url}
\usepackage{latexsym}
\usepackage{xspace}

\pdfoutput=1

\newcommand{\hien}{\textit{hi-en}\xspace}
\newcommand{\bnen}{\textit{bn-en}\xspace}
\newcommand{\teen}{\textit{te-en}\xspace}

\newcommand{\hi}{\textit{hi}\xspace}
\newcommand{\bn}{\textit{bn}\xspace}
\newcommand{\te}{\textit{te}\xspace}
\newcommand{\en}{\textit{en}\xspace}

\aclfinalcopy 

\title{Experiments with POS Tagging Code-mixed Indian Social Media Text}

\author{Prakash B. Pimpale \\
	KBCS, CDAC Mumbai \\
	{\tt prakash@cdac.in} \\\And
	Raj Nath Patel \\
	KBCS, CDAC Mumbai \\
	{\tt rajnathp@cdac.in} \\}

\date{}

\begin{document}
\maketitle
\begin{abstract}
 This paper presents Centre for Development of  Advanced Computing Mumbai's (CDACM)  submission to the NLP Tools Contest on Part-Of-Speech (POS) Tagging For Code-mixed Indian Social Media Text (POSCMISMT) 2015 (collocated with ICON  2015).  We submitted results for Hindi (\hi), Bengali (\bn), and  Telugu (\te) languages mixed with English (\en). In this paper, we have described our approaches to the POS tagging techniques, we exploited for this task. Machine learning has been used to POS tag the mixed language text. For POS tagging, distributed representations of words in vector space (word2vec) for feature extraction and Log-linear models have been tried. We report our work on all three languages \hi, \bn, and \te  mixed with \en.
\end{abstract}


\section{Introduction}
\label{sec:intro}
In this paper, we present our experiments for POS tagging code-mixed Indian social media text. The evolution of social media platforms – such as blogs, micro-blogs (e.g., Twitter), and chats (e.g., Facebook messages) – has created many new sources for information access and language technology. But the same has presented many new challenges, making it
one of the prime present-day research areas.

Most of the Indians and many other Non-English speakers across the world don't always use Unicode to write something in social media, they make use of transliteration and frequently insert English elements through code-mixing and Anglicisms, and often mix multiple languages to express their thoughts. 

English still is the principal language for social media communications, but this kind of multilingual content is growing and calls for the development of language technologies for languages other than English. If we observe twitter and facebook feeds of Indians, it's full of frequent code-mixing. It's not a surprise given the diverse linguistic culture across India. But this poses additional difficulties for automatic Indian social media text processing.

Part-of-speech (POS) is an essential prerequisite for most of the NLP applications. POS tagging of English text are now a quite matured filed in NLP and a lot of work is in progress for English social media text. The work on POS tagging for code-mixed language is a recent topic and not much work has been done for Indian Languages code-mixed text.

~\newcite{vyas:2014} created a multi-level annotated corpus of Hindi-English code-mixed text from facebook forums, and explored language identification, back-transliteration, normalization and POS tagging of this data. They used tools like CRF++ based tagger and Stanford POS tagger for experimentation. ~\cite{jamatia:2015} created a good amount of labeled corpus using amazon mechanical turk and bootstrapping. They experimented with various machine learning techniques for POS tagging and reported Random Forest to be the best one among what they tried.

We have used Stanford log-linear Part-Of-Speech tagger~\cite{toutanova:2000,toutanova:2003} for tagging, word2vec~\cite{Mikolov:MT1:2013} for feature extraction and WEKA~\cite{hall:2009} for machine learning.

The rest of the paper is organized as follows. In section~\ref{sec:data}, we discuss datasets followed by experiments and results in section~\ref{sec:results}. The submission to shared task has been discussed in section~\ref{sec:submit} and the conclusion and future work in section~\ref{sec:con}.

\section{Data-sets}
\label{sec:data}
We have used 80\% of the training data shared by POSCMISMT detailed in Table~\ref{tab:data} for the experiments. Testing for the experiments was done using remaining 20\% data. But the system for final submission was trained using the complete data shared. The submitted systems were evaluated against a test corpus, by the organizers.

\begin{table}
	\centering
	\begin{tabular}{l|c|c|c}
		& & train & test \\ \hline
	\hien & total & 15955 & 11212 \\
		& \hi & 5546 & 411 \\
		& \en & 6178 & 8553 \\
		& O & 4231 & 2248 \\ \hline
	\bnen & total & 24638 & 13561 \\
		& \bn & 8330 & 4671 \\
		& \en & 9973 & 5459 \\
		& O & 6335 & 3431 \\ \hline	
	\teen & total & 4315 & 2255 \\
		& \te & 1716 & 1155 \\
		& \en & 1969 & 819 \\
		& O & 630 &	281
	\end{tabular}
	\caption{Token counts in train and test; O: Others (punctuations, acronyms, named entities, mixed language words and other universal symbols).}
	\label{tab:data}
\end{table}


\section{Experiments and Results}
\label{sec:results}
We have used Stanford POS tagger~\cite{toutanova:2000,toutanova:2003} available on Stanford Natural Language Processing group's website for constrained training and result submission. And unconstrained training and result submission has been done using word2vec~\cite{Mikolov:MT1:2013} and WEKA~\cite{hall:2009}.

\subsection{POS Tagging using Stanford POS tagger: Constrained}
\label{subsec:unconstrained}
The constrained result submission needed to be done using system trained on data provided by POSCMISMT only. We trained Stanford POS tagger using train data provided. Basically this POS tagger learns a log-linear conditional probability model from tagged text, using a maximum entropy method. The POS tag of input word is then decided by the model based on context and surrounding tags of the word. The architecture (arch property) we used for training was: words(-2, 2), order(1), prefix(6), suffix(6), unicodeshapes(1).

\subsection{POS tagging using Machine Learning:	Unconstrained}
\label{subsec:constrained}
We used WEKA to experiment with the application of various machine learning techniques to the POS tagging problem. Various combinations of following word features were used for training and testing the system:
\begin{enumerate}
	\item language of the word
	\item language of the previous word
	\item language of the next word
	\item POS tags of the previous 2 word
	\item POS tags of the next 2 word's similar	words
	\item Position of the word in sentence
\end{enumerate}

In a sentence with length L, words located at positions 1, 2, L and L-1 were assigned required number of default feature values for previous and next languages and POS tags.

For POS tag of the next word's similar word, we used distributed representation of the words in vector space. We trained a word2vec model using train and test corpus detailed in Table~\ref{tab:data}. For the POS tag of the next word, we followed one of the following steps:

\begin{enumerate}
	\item The word was looked up in the list from training data, if it was found, the most frequent POS of that word was used. If it was not in the list, we followed next step.
	\item The nearest word list was fetched using word2vec model trained with the train and test set and the most frequent available POS tag of the nearest word was used instead. If this failed i.e no nearest word was found in the training set, we followed next step.
	\item The most frequent POS tag from the training set was used instead.
\end{enumerate}

\begin{table}
	\centering
	\begin{tabular}{l|c|c}
		& \hien & \teen \\ \hline
	Decision Tree J48 & 44.60 & 50.30 \\
	Decision Tree Random Forest & 43.00 & 47.00 \\
	Naive Bayes & 40.40 & 46.30 \\
	Multilayer Perceptron & 39.30 & 41.10
	\end{tabular}
	\caption{Experimental Results: F1 Measures in \%}
	\label{tab:results}
\end{table}

We reserved 20\% of training data for the purpose of evaluation. Table~\ref{tab:results} details some of the significant results we obtained during the experiments on this test set.


\section{Submission to the Shared Task}
\label{sec:submit}
From the Table~\ref{tab:results} we can see that J48 decision tree gave better results and so that was used to train the final system for submission. The submitted results were evaluated by organizers. These results by organizers have been detailed in Table~\ref{tab:submit} below.

\begin{table}
	\centering
	\begin{tabular}{l|c|c}
		& constrained & unconstrained \\ \hline
	\hien & 71.11 & - \\
	\bnen & 75.46 & - \\
	\teen & 71.04 & 48.03 \\
	\end{tabular}
	\caption{Consolidated Results. Accuracy in \%}
	\label{tab:submit}
\end{table}

\section{Conclusion and Feature Work}
\label{sec:con}
In this paper, we presented two techniques for POS tagging of code-mixed Indian social media text. The method used for constrained submission is performing well, but lack of the quality training data doesn't allow to do much with it. On the other hand, use of the distributed vector representation of words in feature engineering may allow us to use unlabeled
data for training. 

The results are encouraging and future work can be focused on obtaining more social media corpus and using that for the better feature representation.

\bibliography{acl2016}
\bibliographystyle{acl2016}

\end{document}